\documentclass[sigconf]{acmart}

\usepackage{multirow}
\usepackage{booktabs}
\usepackage{tabularx}
\usepackage{array}
\usepackage{enumitem}   
\usepackage{pifont}     
\usepackage{colortbl}
\usepackage[table]{xcolor}

\usepackage{amsmath}
\usepackage{microtype}
\usepackage{listings}
\usepackage{tcolorbox}
\tcbuselibrary{breakable}
\usepackage{url}
\definecolor{errorred}{RGB}{220,53,69}
\definecolor{successgreen}{RGB}{40,167,69}
\definecolor{prologblue}{RGB}{0,102,204}

\AtBeginDocument{%
  }

\copyrightyear{2026}
\acmYear{2026}
\setcopyright{cc}
\setcctype{by}
\acmConference[KDD 2026] {Proceedings of the 32nd ACM SIGKDD Conference on Knowledge Discovery and Data Mining V.2}{August 9--13, 2026}{Jeju Island, Republic of Korea.}
\acmBooktitle{Proceedings of the 32nd ACM SIGKDD Conference on Knowledge Discovery and Data Mining V.2 (KDD 2026), August 9--13, 2026, Jeju Island, Republic of Korea}
\acmISBN{979-8-4007-2259-2/2026/08}
\acmDOI{10.1145/3770855.3818004}

\begin{document}

\title{SymDiag: Explainable Diagnosis for LLM Reasoning via Neuro-Symbolic Verification}

\author{Wenyao Cui}
\orcid{0000-0002-2810-3824}
\affiliation{%
  \institution{Beijing Institute of Technology}
  \city{Beijing}
  \country{China}}
\affiliation{%
  \institution{Zhongguancun Academy}
  \city{Beijing}
  \country{China}}
\email{yao1970099540@gmail.com}

\author{Huaping Zhang}
\orcid{0000-0002-0137-4069}
\affiliation{%
  \institution{Xinjiang Future Enterprise Incubator Co., Ltd.}
  \city{Xinjiang}
  \country{China}}
\affiliation{%
  \institution{Beijing Institute of Technology}
  \city{Beijing}
  \country{China}}
\email{Kevinzhang@bit.edu.cn}

\author{Yongyi Huang}
\orcid{0009-0004-6339-231X}
\affiliation{%
  \institution{Beijing Institute of Technology}
  \city{Beijing}
  \country{China}}
\email{632300420@qq.com}

\author{Qiuchi Li}
\orcid{0000-0002-8219-0869}
\affiliation{%
  \institution{Beijing Institute of Technology}
  \city{Beijing}
  \country{China}}
\email{liqiuchi@bit.edu.cn}

\author{Jian Xu}
\orcid{0009-0001-1090-6207}
\affiliation{%
  \institution{Zhongguancun Academy}
  \city{Beijing}
  \country{China}}
\affiliation{%
  \institution{Institute of Automation, Chinese Academy of Sciences}
  \city{Beijing}
  \country{China}}
\email{jian.xu@ia.ac.cn}

\author{Cheng-Lin Liu}
\orcid{0000-0002-6743-4175}
\affiliation{%
  \institution{Zhongguancun Academy}
  \city{Beijing}
  \country{China}}
\affiliation{%
  \institution{Institute of Automation, Chinese Academy of Sciences}
  \city{Beijing}
  \country{China}}
\email{liucl@nlpr.ia.ac.cn}

\author{Chunxiao Gao}
\orcid{0009-0004-7540-471X}
\affiliation{%
  \institution{Beijing Institute of Technology}
  \city{Beijing}
  \country{China}}
\email{gao_chunxiao@bit.edu.cn}

\author{Juan Wang}
\orcid{0009-0008-4521-3372}
\authornote{Corresponding author.}
\affiliation{%
  \institution{Beijing Institute of Technology}
  \city{Beijing}
  \country{China}}
\email{wangjuan99@bit.edu.cn}

\author{Baohua Zhang}
\orcid{0000-0002-5486-9524}
\affiliation{%
  \institution{Beijing Institute of Technology}
  \city{Beijing}
  \country{China}}
\email{bhzhang_bit@163.com}

\renewcommand{\shortauthors}{Wenyao Cui et al.}

\begin{abstract}
Large language models (LLMs) increasingly serve as data-driven reasoners, yet their chains-of-thought (CoT) can be unfaithful even when final answers are correct. Most existing ``verification'' signals are not diagnostic: answer matching observes only the outcome, LLM-as-judge provides subjective and non-verifiable critiques, and scalar rewards (e.g., PRMs/RMs) offer little insight into where a multi-step derivation fails.We propose \textbf{SymDiag}, a neuro-symbolic framework that \textbf{reframes reasoning verification as structured failure diagnosis}. SymDiag translates natural-language CoT into symbolic constraints and performs step-level satisfiability/entailment checks to (i) localize failing steps and (ii) produce verifiable diagnostic evidence, including counterexamples, inconsistency witnesses, and missing-premise indicators. A central challenge is that apparent ``logic violations'' can be caused either by genuine reasoning defects or by neural-to-symbolic translation noise. SymDiag therefore incorporates a Self-Auditor that disentangles TranslationError from ReasoningError via dual symbolic encodings consistency checks, enabling robust diagnosis under partial observability. Across diverse mathematical, logical, scientific, and general reasoning benchmarks, SymDiag improves detection of unfaithful reasoning and provides substantially more effective feedback for multi-round reasoning repair than outcome-only verification and LLM-based judging, offering a principled foundation for trustworthy and scalable reasoning diagnosis.
\end{abstract}



\maketitle

\section{Introduction}

Large language models (LLMs) exhibit strong multi-step reasoning~\cite{kalyanpur2024multi,cai2025role,zhao2024large}, and have rapidly been adopted across a wide range of downstream domains---from chain-of-thought-driven perception and controllable image editing~\cite{li2024image,cai2025bayesian}, motion understanding, generation, and repetitive action counting~\cite{li2026multiple,li2025human,yao2025countllm,gu2025mocount,jia2026ram}, multimodal speech and medical analysis~\cite{yang2025you,tong2025pamn,shi2025medal,xu2026robust}, to controllable 3D content generation and reconstruction~\cite{yan2026scene,liu2024graph,zhang2026psgs,chen2025dense}; concurrent work also characterizes their scaling, long-context, and few-shot learning behavior~\cite{shi2024scaling,shi2026intrinsic,guan2025meta}. Yet their chains-of-thought (CoT) can be unfaithful: intermediate steps may be inconsistent, rely on hidden assumptions, or make non-entailing transformations even when the final answer is correct. This gap between outcome correctness and process validity is a key barrier to deploying LLMs in high-stakes settings \cite{yang2025neurosurvey,wang2024towards,qiu2025quantifying}.

Most existing ``verification'' signals are not designed for diagnosis. Answer matching (outcome-only evaluation) cannot localize failures in a long derivation; LLM-as-judge produces natural-language critiques that are subjective and hard to audit; and reward models/process reward models reduce rich failure modes to a scalar score, providing little guidance for targeted repair. Recent work has explored stronger supervision, including logic-aware rewards \cite{logicreward2025}, formally verified process labels for process reward models \cite{fover2026}, information-theoretic step-level optimization for self-refining LLMs~\cite{zhao2026beyond}, and additional logic training on synthetic corpora \cite{alt-logic-training}. While these advances improve accuracy or faithfulness on average, they largely treat verification as scoring rather than explaining where and why reasoning fails.

Neuro-symbolic methods offer a path toward auditable reasoning by introducing explicit symbolic structure \cite{chen2025comparative}. Logic-LM translates natural language into logic and delegates inference to solvers \cite{logic-lm}, and LogicReward constructs logic-aware reward signals grounded in formal checks \cite{logicreward2025}. SymbCoT maintains symbolic expressions alongside natural language \cite{symbcot}, Aristotle introduces a logic-complete decompose-search-resolve framework \cite{Aristotle25}, and Symbol-LLM develops unified text-to-symbol training \cite{symbol-llm,quan2025peirce,sadowski2025explainable,hersche2024towards}. However, even when symbolic backends are used, a central practical challenge remains: failures may stem from either \textbf{genuine reasoning defects} or \textbf{neural-to-symbolic translation artifacts}. Without separating these sources, a verifier can ``misdiagnose'' parsing/formalization noise as logical error, producing unreliable feedback.

\begin{figure}
    \centering
    \includegraphics[width=1\linewidth]{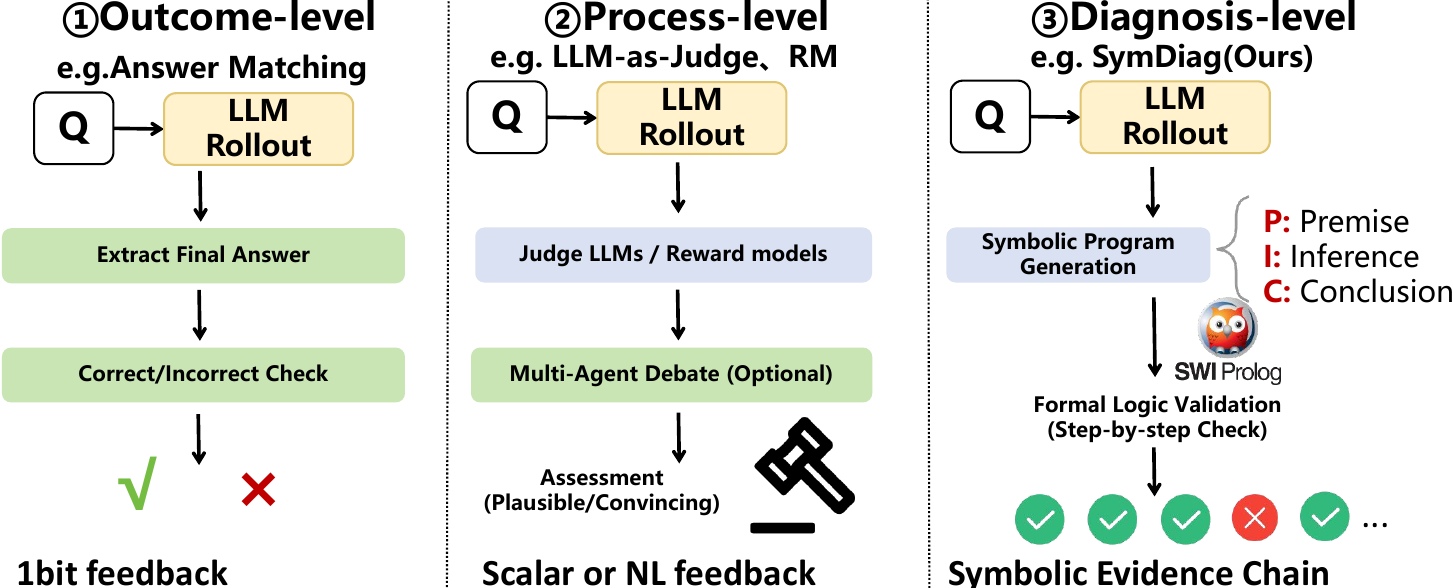}
    \caption{Paradigms for evaluating chain-of-thought (CoT) reasoning. Outcome- and process-level approaches treat verification as scoring, judging only final answers or subjective plausibility. SymDiag introduces a diagnosis-level paradigm, performing symbolic checks to localize reasoning failures and produce verifiable diagnostic evidence, enabling targeted repair.}
    \label{fig:method_compare}
\end{figure}

We propose \textbf{SymDiag}, an explainable neuro-symbolic framework that \textbf{reframes reasoning verification as structured failure diagnosis}. SymDiag performs step-level symbolic checks over CoT, localizes failure points, and produces verifiable evidence (e.g., counterexamples, inconsistency witnesses, and missing-premise indicators) that is actionable for repair.

To make diagnosis reliable under noisy or underspecified CoT, SymDiag incorporates a Self-Auditor that explicitly disentangles TranslationError from ReasoningError. The Self-Auditor generates dual symbolic encodings of each natural-language step and applies cross-branch consistency and lightweight sanity checks; when an apparent violation disappears under minimal text-consistent reformulations, it is attributed to translation rather than reasoning. This design is crucial for trustworthy diagnosis under partial observability.

Another line of work integrates theorem provers to verify and refine natural language explanations \cite{explanation-refiner,quan2024enhancing}. LogicReward and FoVer similarly rely on formal verification to create training signals \cite{logicreward2025, fover2026}. SymDiag differs in goal and output: instead of returning only a score or a corrected trace, it localizes failures and returns checkable diagnostic evidence that attributes what went wrong and supports targeted repair across domains.

\begin{tcolorbox}[
    title=Takeaway,
    colback=blue!5,
    colframe=black!60,
    boxrule=0.8pt,
    arc=3pt,
    left=6pt,
    right=6pt,
    top=6pt,
    bottom=6pt
]
\textbf{SymDiag is not a better verifier or a better reward for reasoning, but an explainable neuro-symbolic diagnostic system that localizes, attributes, and repairs reasoning failures with verifiable evidence across domains.}
\end{tcolorbox}

Our main contributions are:

\textbf{(1) A diagnosis-centric view of LLM reasoning failures.}
We move beyond outcome- or reward-based verification and formalize reasoning verification as a \textbf{failure diagnosis} problem, aiming to localize errors, attribute their causes, and support targeted repair.

\textbf{(2) Self-audited diagnosis that disentangles translation errors from genuine reasoning failures.}
We introduce a self-auditing mechanism that explicitly distinguishes errors arising from neural-to-symbolic translation artifacts from true logical reasoning defects. By leveraging dual symbolic encodings and cross-branch consistency checks, the Self-Auditor prevents spurious misdiagnosis caused by underspecified or ambiguous natural-language translations, enabling reliable step-level verification and diagnosis under partial observability.

\textbf{(3) An evidence-grounded diagnosis-and-repair pipeline.}
We propose SymDiag, a novel neuro-symbolic framework that performs step-level symbolic verification, attributes failures with verifiable evidence (e.g., counterexamples, unsat cores, missing premises), and guides targeted repair. This closed-loop process significantly improves multi-round reasoning correction.

\textbf{(4) A cross-domain diagnostic benchmark and dataset.}
We construct a unified diagnostic dataset spanning mathematics (e.g., AIME), logical reasoning(e.g., AR-LSAT), scientific reasoning (e.g., GPQA), and general reasoning (e.g., MMLU), demonstrating that diverse reasoning errors can be analyzed within a shared symbolic diagnostic framework.

\section{Related Work}

We review prior work through the lens of what kind of supervision or feedback signal it produces (outcome labels, natural-language critiques, scalar rewards, or formal evidence), and highlight why existing approaches are often insufficient for \textbf{failure diagnosis}.

\subsection{Outcome-Only Verification and LLM-as-Judge}

The most common verification protocol evaluates only final answers against gold labels \cite{cobbe2021training}. While efficient, outcome-only evaluation cannot detect unfaithful-but-correct traces and cannot localize where errors occur in multi-step derivations.

LLM-as-Judge methods extend outcome checking by prompting language models to assess solution quality and provide critiques \cite{llm-as-judge1,llm-as-judge2,cao2025pretraining}. However, such judgments are inherently subjective and inconsistent, and may prefer fluent but logically flawed reasoning \cite{zheng2023judging}; even attention-based explanations of LLM decisions provide only coarse, non-verifiable signals at the step level~\cite{lan2025attention}. As a result, the feedback is difficult to audit and hard to convert into reliable, step-local repair actions.

\subsection{Scalar Rewards and Process Supervision}

To provide denser training or selection signals, process reward models (PRMs) score intermediate reasoning steps \cite{step_reward,step_reward2,lightman2023let,uesato2022solving}. PRMs can improve best-of-$n$ sampling and overall faithfulness, but they fundamentally treat verification as scoring: the output is typically a scalar reward rather than a structured explanation of what violated which constraint.

Recent work explores scalable labeling and training signals, including synthetic error injection and solver-based labeling \cite{prm1,prm2,fover2026}. Logic-aware reward constructions similarly aim to align models with formally checked objectives \cite{logicreward2025}. Parameter-efficient fine-tuning further enables scalable training of verifiers and judges under tight compute budgets~\cite{zhao2025tiny,zhang2025uora}. Despite stronger signals, these approaches generally do not provide (i) explicit step-level localization, (ii) evidence that can be independently checked, or (iii) an attribution that separates reasoning defects from translation/formatting artifacts.

\subsection{Neuro-Symbolic Reasoning and Symbolic Interfaces}

A growing body of work integrates symbolic structure to strengthen LLM reasoning, complementing analyses of how LLMs perform deductive and inductive inference~\cite{cai2025role}. Logic-LM translates natural language into formal logic and delegates inference to solvers \cite{logic-lm}; LogicReward constructs logic-aware reward signals grounded in formal checks \cite{logicreward2025}; LINC uses first-order logic as an intermediate representation \cite{linc}; SymbCoT maintains explicit logical expressions alongside natural language \cite{symbcot}; Aristotle introduces a decompose-search-resolve framework with embedded symbolic rules \cite{Aristotle25}; and Symbol-LLM develops unified text-to-symbol training and tuning \cite{symbol-llm}.

Logic-programming backends have also been used to improve reliability, e.g., combining LLMs with a Prolog interpreter to generate reasoning proofs \cite{yang2025neuro}. Related neuro-symbolic frameworks address specialized settings such as temporal abductive reasoning \cite{liang2025nestr} or ontological reasoning \cite{vsevolodovna2025enhancing}. These systems demonstrate the value of explicit symbolic structure, but they primarily focus on solving tasks (or producing proofs) rather than producing diagnostic outputs that explain and localize failures.

\subsection{Formal Verification and Neuro-Symbolic Verification}

Theorem provers and formal tools can provide machine-checkable guarantees beyond neural evaluation. LogicReward uses provers to generate step-level logical rewards \cite{logicreward2025}, and FoVer applies formal verification to label reasoning steps for PRMs \cite{fover2026}. Explanation-Refiner formalizes NLI explanations and provides fine-grained prover feedback for iterative correction \cite{explanation-refiner}. Sultan et al.~\cite{sultan2025towards} study reliable proof generation with neuro-symbolic components.

In mathematics, Lean-based systems \cite{yang2023leandojo} and hybrid provers \cite{hu2025hybridprover} provide strong guarantees but often require domain-specific libraries. Tool-integrated approaches can verify computation via interpreters or calculators \cite{feng2025retool}, but do not directly address logical entailment in general reasoning.

Several concurrent works explore neuro-symbolic verification gates and verifier-in-the-loop reasoning, e.g., Eidoku \cite{miya2025eidoku}, VIRO \cite{park2026viro}, and neuro-symbolic training-time consistency objectives \cite{calanzone2024logically}. \textbf{Despite these advances, a common limitation persists: verification often returns a score or a corrected output without explicitly localizing error sources, providing checkable evidence, or disentangling translation artifacts from genuine reasoning defects.}

\textbf{SymDiag} addresses this gap by reframing verification as \textbf{failure diagnosis}: it localizes failing steps, attributes failure types, and returns verifiable evidence (counterexamples, inconsistency witnesses, missing-premise indicators) together with a self-audited attribution of TranslationError vs. ReasoningError, enabling targeted repair across diverse reasoning domains.

\begin{figure*}
    \centering
    \includegraphics[width=1\linewidth]{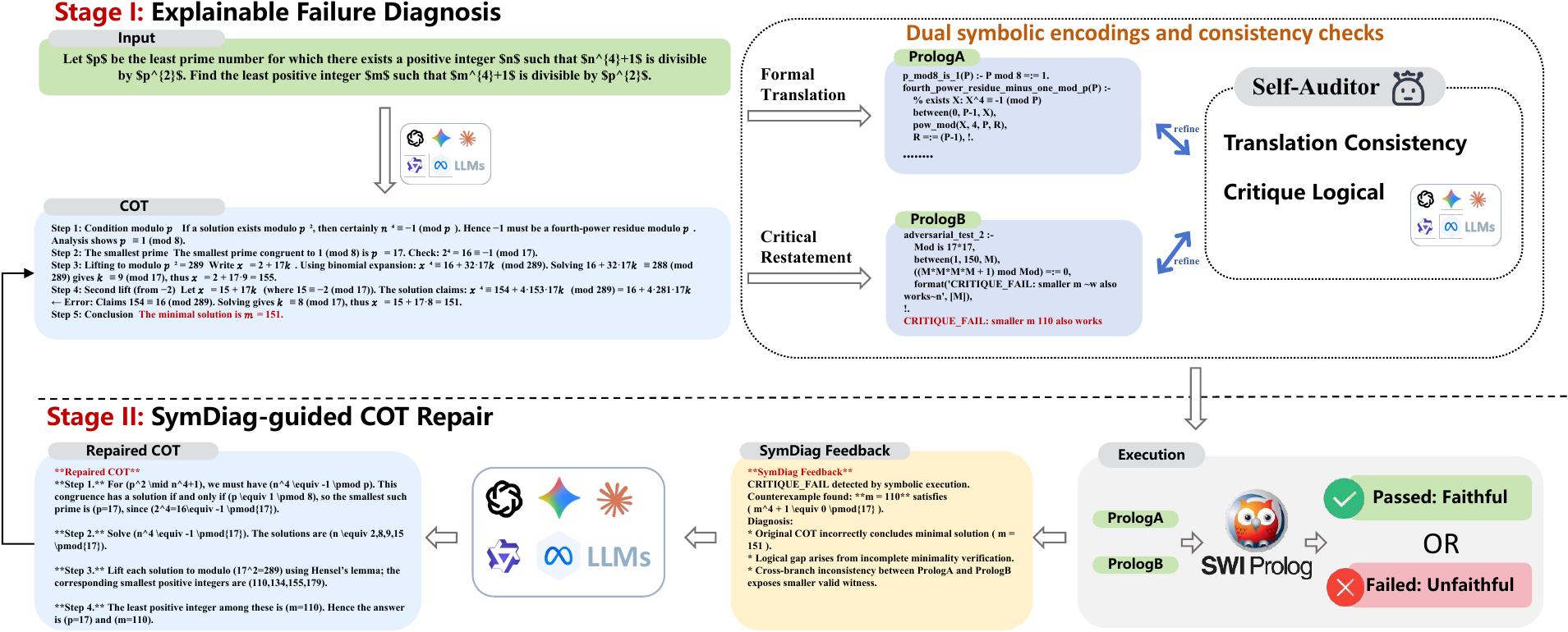}
    \caption{SymDiag overview. \textbf{Stage~I (Diagnosis):} a neuro-symbolic generator produces (i) a formal translation and (ii) a critical restatement of the original CoT as two independent Prolog programs; a Self-Auditor checks cross-encoding consistency to distinguish TranslationError from ReasoningError, and SWI-Prolog performs step-level satisfiability/consistency checks to output a faithful/unfaithful decision with verifiable evidence (e.g., counterexamples, inconsistency witnesses, missing-premise indicators). \textbf{Stage~II (Repair):} SymDiag uses localized failures and evidence to prompt an LLM to generate a repaired reasoning trace that is solver-consistent.}
    \label{fig:SymDiag}
\end{figure*}

\section{Methodology}
\label{sec:methodology}

SymDiag is a two-stage neuro-symbolic framework for \textbf{explainable failure diagnosis} and diagnosis-guided repair of LLM reasoning traces (Figure~\ref{fig:SymDiag}). Unlike outcome-only checking or natural-language critique, SymDiag (i) compiles natural-language reasoning into verifiable symbolic constraints, (ii) performs step-level satisfiability/entailment validation, (iii) localizes failing steps and attributes failure types with \textbf{checkable evidence}, and (iv) converts the diagnosis into actionable feedback for iterative repair.

\subsection{Problem Setup and Outputs}
\label{sec:problem_setup}

Given an input problem $x$ (question + optional context) and an LLM-produced chain-of-thought (CoT)
$y_{\text{cot}}=(s_1,\ldots,s_T)$ with final answer $\hat{y}$, our goal is to assess the \emph{faithfulness} of $y_{\text{cot}}$ and diagnose failures. SymDiag outputs:

\begin{itemize}
    \item \textbf{Step-level verdicts:} $v_i \in \{\texttt{pass},\texttt{fail}\}$ for each step $s_i$;
    \item \textbf{Localization:} a set of failing indices $\mathcal{F}\subseteq\{1,\ldots,T\}$;
    \item \textbf{Error labels:} $\ell_i$ from a predefined taxonomy;
    \item \textbf{Verifiable evidence:} symbolic artifacts $\mathcal{E}_i$ such as counterexamples, inconsistency witnesses (unsat cores), and missing-premise indicators.
\end{itemize}

We emphasize that SymDiag diagnoses can flag \emph{unfaithful-but-correct} reasoning traces (correct $\hat{y}$ but invalid intermediate entailments), which answer-based evaluation cannot detect.

\subsection{Stage I: Neuro-Symbolic Compilation and Verification}
\label{sec:stage1}

Stage~I transforms the natural-language CoT into a symbolic representation and verifies it step-by-step.

\subsubsection{Step State Representation}
\label{sec:state_repr}

We represent the reasoning process as a sequence of symbolic states:
$$
S_i = \{P_i, I_i, C_i\},
$$
where (i) $P_i$ encodes accumulated \textbf{premises} and derived facts up to step $i$,
(ii) $I_i$ encodes the \textbf{intended inference} performed in step $s_i$ (e.g., a rule application, algebraic transformation, or entailment claim),
and (iii) $C_i$ encodes \textbf{constraints} (domain restrictions, type constraints, boundary conditions, and task-specific axioms).
Each $S_i$ is compiled into a symbolic program fragment in a solver-executable form (we use Prolog as the backend\footnote{We adopt Prolog as the backend because it natively supports unification, backtracking, and Horn-clause resolution while remaining lightweight and human-readable, unlike heavier proof assistants (e.g., Lean, Coq) that require domain-specific formal libraries.}).

\subsubsection{Neuro-Symbolic Generator}
\label{sec:generator}

Natural-language reasoning is often underspecified. To mitigate brittleness in translation, we use a \textbf{two-branch generator} that produces two independent symbolic encodings:

\paragraph{(A) Formal Translation Branch.}
This branch compiles $x$ and $y_{\text{cot}}$ into a formal representation by mapping entities, predicates, and relations into a consistent signature. For math/science tasks, it additionally normalizes quantities, units, and equalities/inequalities; for logic tasks, it explicitly models quantifiers and scope.

\paragraph{(B) Critical Restatement Branch.}
This branch re-expresses the CoT step $s_i$ into a stricter, explicitly-scoped statement, then compiles the restated claim into symbolic form. The intent is to expose hidden assumptions that the surface text may conceal.

Let the resulting programs be $\Pi^{(A)}$ and $\Pi^{(B)}$, each providing step-level states $\{S_i^{(A)}\}$ and $\{S_i^{(B)}\}$. \textbf{The dual encodings reduce single-path translation bias and increase robustness to paraphrase variability.}

\subsubsection{Syntax Check and Normalization}
\label{sec:syntax}

Before logical verification, we perform deterministic checks:
(i) signature consistency (arity/type sanity),
(ii) groundability checks (detecting free variables that should be bound),
(iii) constraint normalization (canonicalizing equalities and domain constraints),
and (iv) solver-compatibility checks. If a program fails syntax validation, it triggers a TranslationError pathway (Section~\ref{sec:self_auditor}).

\subsubsection{Self-Auditor: Translation vs. Reasoning Error Disentanglement}
\label{sec:self_auditor}

A key design goal is to separate genuine reasoning failures from translation artifacts. The Self-Auditor consists of two tests:

\paragraph{Translation Consistency Check.}
We compare $\Pi^{(A)}$ and $\Pi^{(B)}$ at the level of extracted facts, constraints, and entailment targets. Large divergences suggest translation ambiguity or mapping failure. Concretely, we compute overlap statistics between fact sets and constraints, and we flag steps where the two branches imply incompatible symbolic targets for the same natural-language claim.

\paragraph{Logical Critique Check.}
We run lightweight symbolic sanity tests (e.g., immediate contradictions, impossible type assignments, constraint violations) to detect whether an apparent failure is an encoding artifact. If the failure disappears under minimal canonical rewrites (e.g., variable renaming, constraint relaxation consistent with the text), we attribute it to translation.

The Self-Auditor outputs a binary attribution:
$$
a \in \{TranslationError, ReasoningError\},
$$
and passes only \texttt{Approved} states to the step-level verifier.

\subsubsection{Step-Level Symbolic Verification}
\label{sec:step_verification}

For each step state $S_i$, we verify whether the step’s inference is supported by prior information. We implement two complementary checks:

\paragraph{(1) Consistency / Satisfiability.}
We test whether $P_i \cup C_i$ is satisfiable:
$$
\mathrm{SAT}(P_i \wedge C_i).
$$
If unsatisfiable, the step has introduced an inconsistency (often due to contradictory assumptions, invalid algebraic manipulation, or misapplied rule constraints).

\paragraph{(2) Local Entailment of the Step Claim.}
Let $\varphi_i$ be the symbolic claim corresponding to step $s_i$ (encoded in $I_i$). We test:
$$
(P_{i-1} \wedge C_{i-1}) \models \varphi_i,
$$
operationalized by checking unsatisfiability of the negation:
$$
\mathrm{UNSAT}(P_{i-1} \wedge C_{i-1} \wedge \neg \varphi_i).
$$
If satisfiable, we can produce a \textbf{counterexample} assignment that falsifies $\varphi_i$ while respecting prior premises.

\paragraph{Faithfulness Decision.}
A step is marked \textbf{fail} if it introduces inconsistency or if the claimed entailment does not hold. The full CoT is \textbf{faithful} only if all steps pass in at least one approved branch, with ties broken conservatively (we prefer \textbf{unfaithful} when both branches fail under auditing).

\subsubsection{Symbolic Diagnosis: Failure Localization and Evidence}
\label{sec:diagnosis}

When verification fails, SymDiag generates a diagnosis tuple for each failing step $i\in\mathcal{F}$:
$$
d_i = (\ell_i, \mathcal{E}_i, \text{scope}_i),
$$
where $\ell_i$ is an error label, $\mathcal{E}_i$ is evidence, and $\text{scope}_i$ indicates whether repair should be local (patch step) or global (rewrite reasoning prefix/suffix).

\subsection{Stage II: Diagnosis-Guided Repair}
\label{sec:stage2}

Stage~II converts the symbolic diagnosis $\{d_i\}$ into actionable feedback and uses it to repair the original reasoning trace.
For each failing step $i\in\mathcal{F}$, we verbalize the error label $\ell_i$ and attach checkable evidence $\mathcal{E}_i$ (e.g., counterexample assignments, inconsistency witnesses, or missing-premise indicators) to form a structured feedback message.

\paragraph{Local patch vs. global rewrite.}
If $\text{scope}_i$ indicates a localized defect (e.g., arithmetic/algebra error, invalid equivalence in one step), we prompt the base model to minimally edit the offending step while preserving the surrounding context.
If $\text{scope}_i$ indicates that the failure propagates (e.g., early missing premise, type mismatch affecting later steps), we prompt the model to rewrite the reasoning from the earliest failing step onward.

\subsection{Dataset Construction}
\label{sec:data_construction}
\begin{figure}[htbp]
    \centering
    \includegraphics[width=1\linewidth]{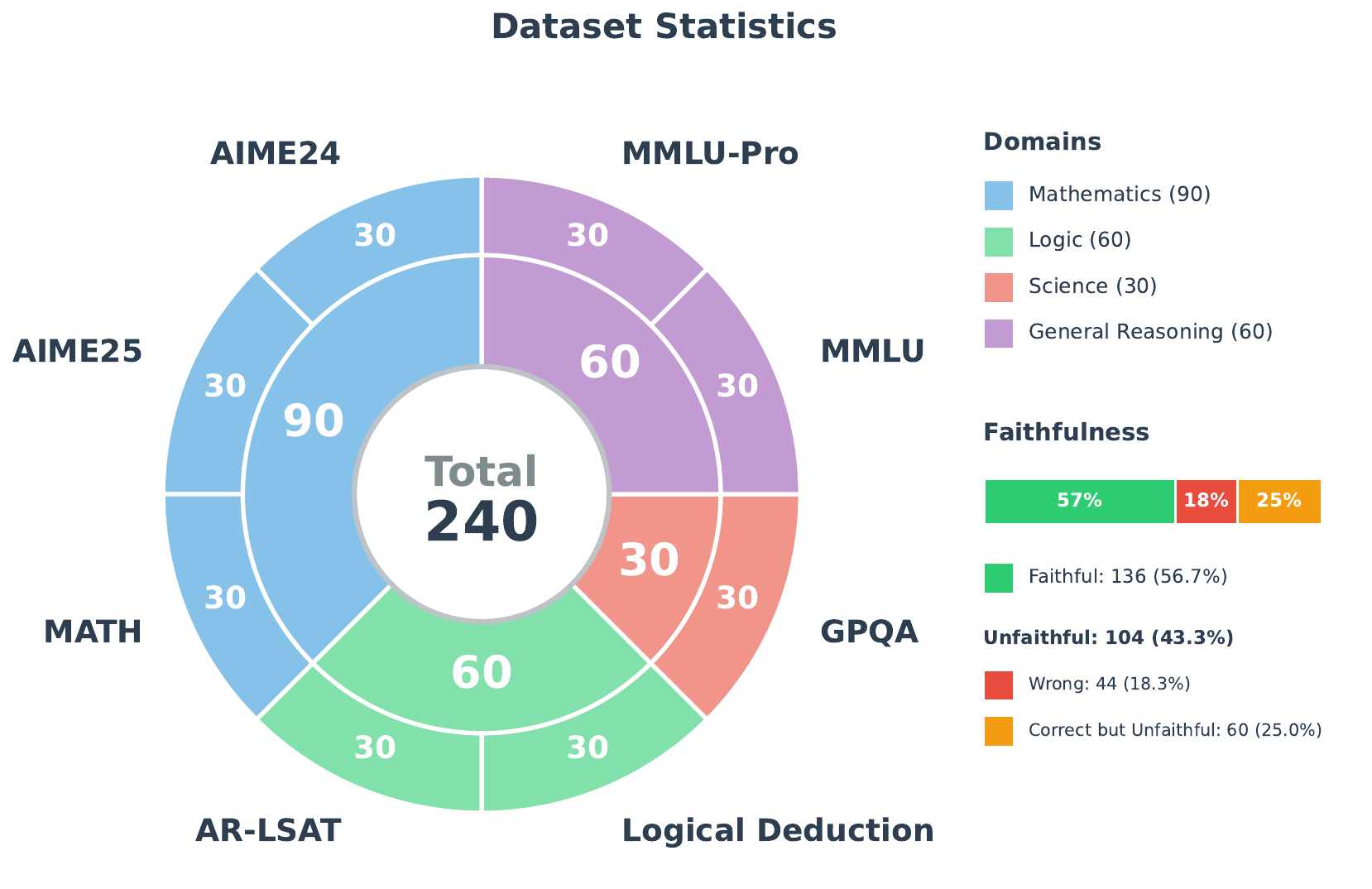}
\caption{Core experimental dataset composition. We manually audit 240 instances in total, sampling 30 examples from each dataset across four reasoning domains.}
\label{tab:dataset_stats}
\end{figure}

We construct a unified diagnostic dataset for CoT faithfulness analysis, where each instance is assigned one of two high-level reasoning-trace labels: (i) \textbf{Faithful} (correct final answer and no verifiable defect in the CoT) and (ii) \textbf{Unfaithful} (a clear, locatable defect, including missing premises, invalid equivalences, ignored boundary conditions, or defects that lead to an incorrect final answer). 

Briefly, we build a large-scale corpus through a two-stage pipeline. \textbf{Stage~1} samples CoTs and final answers from base LLMs (Section~\ref{sec:models}); incorrect-answer instances are directly labeled as \textbf{Unfaithful}. \textbf{Stage~2} applies conservative multi-judge voting on correct-answer instances to retain high-precision \textbf{Faithful} traces.

\textbf{Gold set for evaluation.}
While the full automatically constructed corpus contains \textbf{437,792} instances, our main experiments use a manually verified \emph{gold set} of \textbf{240} instances sampled from this corpus to enable reliable, human-audited evaluation (Table~\ref{tab:dataset_stats}).

\section{Experiments}
\subsection{Experiments Setup}
\subsubsection{Datasets.}
All core experiments are conducted on our manually audited diagnostic dataset (Section~\ref{sec:data_construction}, Table~\ref{tab:dataset_stats}, a curated subset of 240 instances spanning four domains: \textbf{mathematics}, \textbf{logical}, \textbf{science}, and \textbf{general reasoning}. Each instance is annotated with a reasoning-trace label (\textbf{Faithful} or \textbf{Unfaithful}).

\subsubsection{Baselines.}
To evaluate the effectiveness of SymDiag, we compare it against representative baselines that rely on weaker or non-verifiable supervision signals. The comparison focuses on two aspects: (i) \textbf{Diagnosis}: detection of Faithful versus Unfaithful reasoning traces, and (ii) \textbf{Feedback and Repair}: the effectiveness of each method's feedback in guiding iterative reasoning repair. We consider the following baselines:

\begin{itemize}
    \item \textbf{Answer Matching.} Faithfulness is judged solely by exact matching of the final answer. During repair, the model is provided with a 1-bit feedback signal indicating whether the final answer is correct.
    \item \textbf{LLM-as-Judge.} A judge LLM provides natural-language feedback by evaluating both the final answer and the chain-of-thought (CoT), after which the model reattempts the solution accordingly. For fair comparison, both this baseline and our method use \textbf{GPTOSS-120B} as the base model, and we adopt multi-agent aggregation (i.e., multiple independent judges with voting) to keep the overall inference budget comparable.
    \item \textbf{Reward Model.} A scalar score is returned as feedback. We use a reward model from NVIDIA~\cite{wang2025helpsteer3preferenceopenhumanannotatedpreference} to score candidate solutions and provide the resulting reward signal for reattempt or repair.\footnote{\url{https://huggingface.co/nvidia/Qwen-3-Nemotron-32B-Reward}}
    \item \textbf{LogicReward.} LogicReward returns a scalar-valued signal, and the underlying Isabelle prover outputs (e.g., proof states / error messages) as feedback. We adopt theorem-prover-based logical rewards as a verification signal for reasoning steps~\cite{logicreward2025}\footnote{This method uses Isabelle rather than Prolog.}. For fair comparison, both this baseline and our method use \textbf{GPTOSS-120B} as the base model.
\end{itemize}

\subsubsection{Models.} 
\label{sec:models}
We evaluate SymDiag in the context of diagnosis-guided reasoning repair. The models under study serve as base reasoners whose chain-of-thoughts are verified, diagnosed, and repaired by SymDiag, rather than as verifiers themselves. To assess robustness across model capacity, we consider both small and large open-weight instruction-tuned LLMs: Llama-3.2-1B~\cite{grattafiori2024llama}, Qwen3-1.7B~\cite{yang2025qwen3}, Qwen3-8B~\cite{yang2025qwen3}, GPTOSS-20B~\cite{agarwal2025gpt}. For fairness, all baselines that require an LLM-based verifier/judge/feedback generator (including LLM-as-Judge and LogicReward) use the same judge model GPTOSS-120B~\cite{agarwal2025gpt}; the only exception is the Reward Model baseline, which uses Qwen-3-Nemotron-32B-Reward\footnote{\url{https://huggingface.co/nvidia/Qwen-3-Nemotron-32B-Reward}}. Unless otherwise specified, the diagnosis, error attribution, and feedback generation components of SymDiag are implemented using a stronger judge model GPTOSS-120B~\cite{agarwal2025gpt}, ensuring stable reasoning, consistent symbolic translation, and reliable diagnostic feedback.

\subsubsection{SymDiag Configuration.}
For each example, the base LLM produces a chain-of-thought $y_{\text{cot}}$. The Neuro-Symbolic Generator translates and critiques the CoT into a sequence of symbolic step records, which are verified step-by-step by the symbolic solver backend. If verification fails, SymDiag performs two-branch diagnosis (TranslationError vs ReasoningError), localizes failing steps, attaches an error label, and returns evidence-grounded feedback to drive either local patching or full-chain rewriting. We run up to $N$ repair rounds (default $N{=}4$, unless otherwise stated) and report both single-round and multi-round outcomes.

\subsubsection{Metrics.}
We evaluate \textbf{SymDiag} along two complementary dimensions: (i) its ability to distinguish \textbf{Faithful} from \textbf{Unfaithful} reasoning traces, and (ii) the effectiveness of diagnosis-driven feedback in supporting iterative reasoning repair.

\textbf{Faithfulness Detection.}
We treat each method as a binary classifier that predicts whether a given reasoning trace is \texttt{Faithful} or \texttt{Unfaithful}. For this task, we report the F1 score of faithfulness detection on each dataset, with detailed results summarized in Table~\ref{tab:results_main}.

\textbf{Repair Effectiveness.}
To assess how informative and effective different feedback signals are for reasoning repair, we evaluate task performance after each repair round using standard benchmark metrics. We report per-dataset scores and learning curves, enabling a direct comparison of how different methods influence both the effectiveness and sample efficiency of multi-round reasoning correction (see Figure~\ref{fig:repair_curves_all}).

\begin{table*}[htbp]
    \centering
    \includegraphics[width=\textwidth]{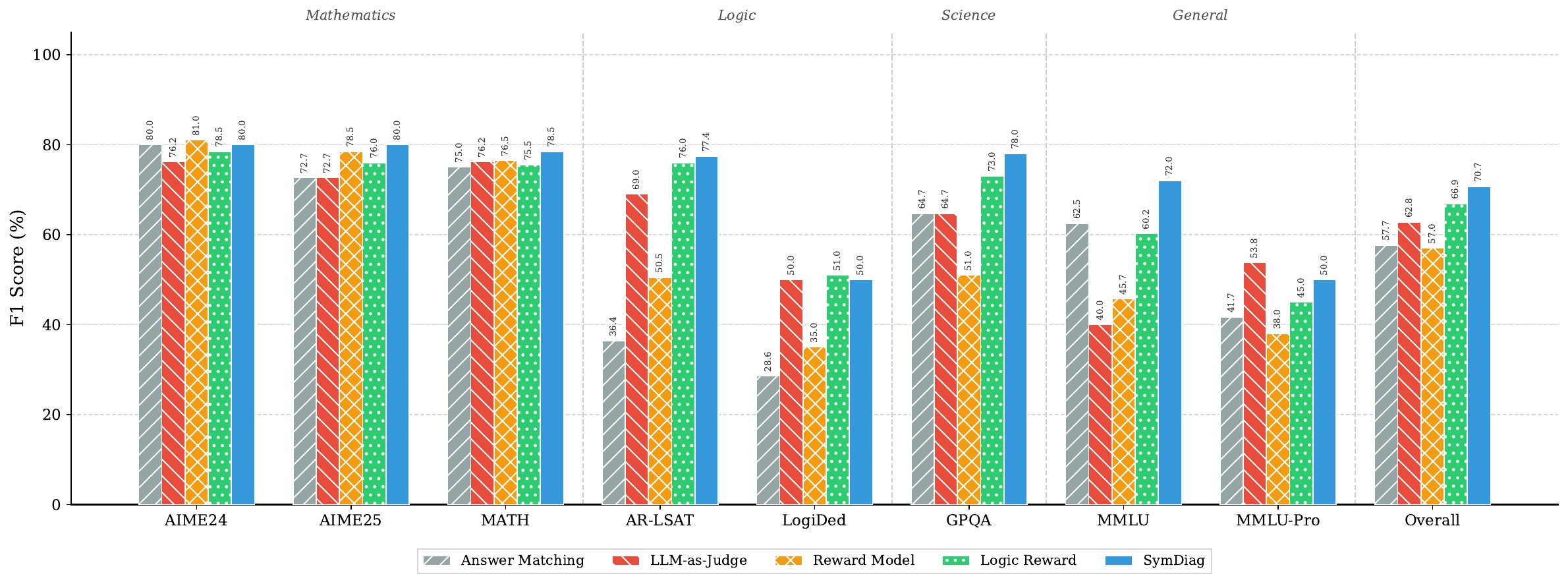}
    \caption{Faithfulness detection performance (F1) across datasets and overall. All results are evaluated on the manually audited diagnostic subset described in Section~\ref{sec:data_construction}. The \textit{Overall} score is computed over all instances.}
    \label{tab:results_main}
\end{table*}

\section{Results}
\label{sec:results}

We present experimental results addressing two core questions:

\begin{tcolorbox}[
    title=Research Questions,
    colback=blue!5,
    colframe=black!60,
    boxrule=0.8pt,
    arc=3pt,
    left=6pt,
    right=6pt,
    top=6pt,
    bottom=6pt
]
\textbf{(Q1)} Can \textbf{SymDiag} more accurately detect \textbf{Unfaithful} reasoning traces?

\textbf{(Q2)} Does \textbf{SymDiag} provide more effective feedback for iterative reasoning repair?

\end{tcolorbox}

All results are reported on the manually audited diagnostic dataset of 240 instances described in Section~\ref{sec:data_construction}.

\subsection{Faithfulness Detection Performance}
\textbf{SymDiag is significantly more effective at identifying unfaithful reasoning.} As shown in Table~\ref{tab:results_main}, SymDiag achieves the best faithfulness detection performance across all datasets, with the highest overall F1 score (70.7). It consistently outperforms answer matching, LLM-as-Judge, and reward-based baselines, demonstrating that outcome-only or scalar-reward signals are insufficient for identifying unfaithful reasoning. The performance gap is especially large on logical and general reasoning benchmarks (e.g., AR-LSAT, LogiDed, MMLU), where correct answers are often produced via invalid intermediate steps that baselines fail to detect. Even in mathematics, where baselines perform relatively well, SymDiag provides consistent gains by localizing subtle step-level violations. These results highlight the core advantage of SymDiag: by performing symbolic, step-level diagnosis with verifiable evidence, it detects unfaithful chains-of-thought more reliably and robustly than heuristic judging or reward-based verification across diverse reasoning domains.

\subsection{Diagnosis-Guided Reasoning Repair}
\label{sec:repair}

\begin{figure*}[htbp]
    \centering
    \includegraphics[width=1.0\linewidth]{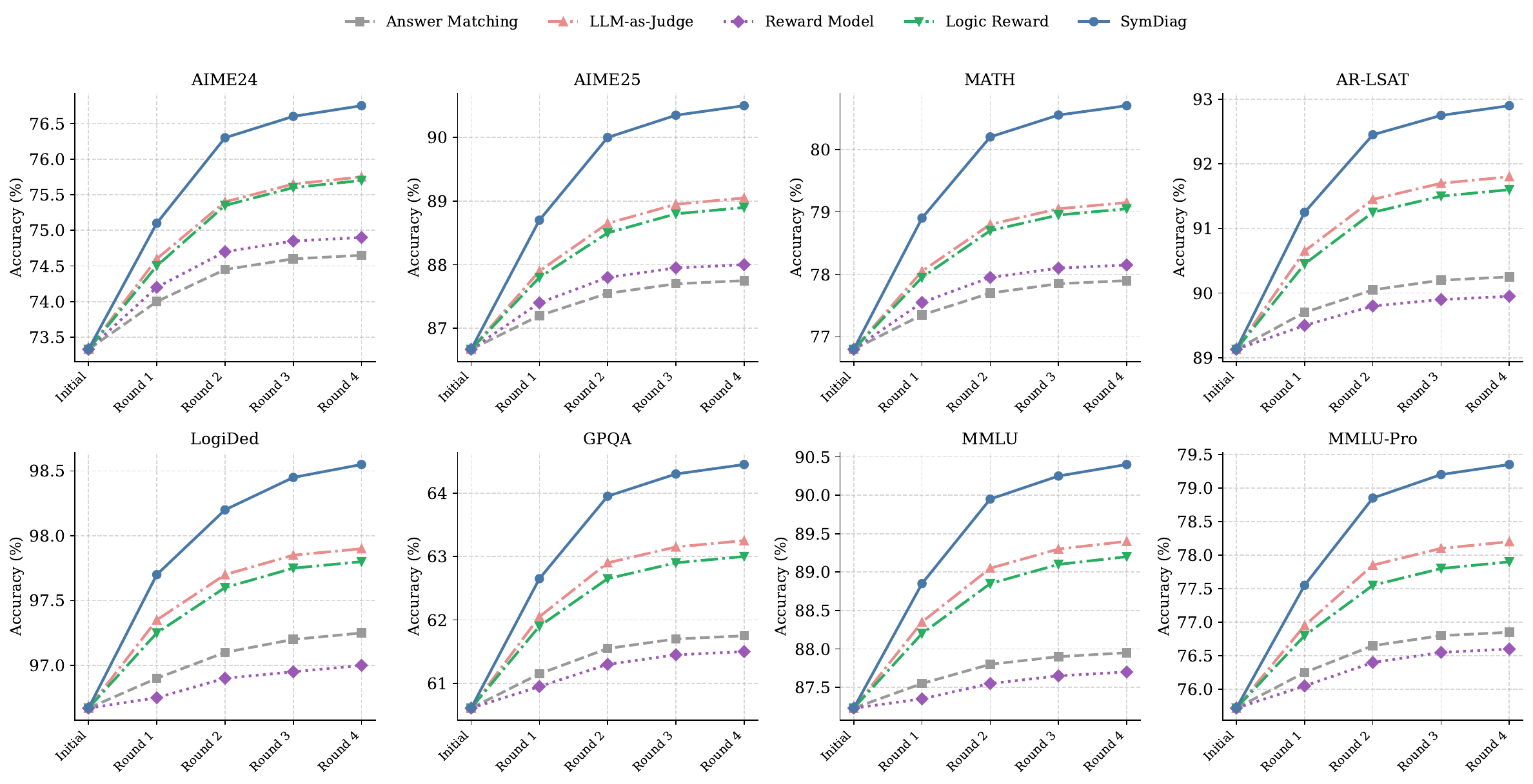}
    \caption{Diagnosis-guided reasoning repair curves across datasets. Each subplot reports task accuracy after each repair round (Round~0 is the original answer). SymDiag yields faster and more sustained gains, reflecting the benefit of localized, verifiable error evidence for targeted correction.}
    \label{fig:repair_curves_all}
\end{figure*}

\textbf{SymDiag can more effectively guide the correction of reasoning errors.} Figure~\ref{fig:repair_curves_all} shows that SymDiag consistently achieves faster and larger gains across repair rounds than all baselines. Answer Matching yields minimal improvement due to the lack of localized guidance, while Reward Model and LogicReward provide scalar signals that are not explicitly localizable and can be noisy, especially in open-domain settings. LLM-as-Judge offers moderate early gains but quickly saturates because its feedback is coarse and non-verifiable. In contrast, SymDiag provides step-level, evidence-grounded diagnoses that directly target invalid inferences, missing premises, and constraint violations, enabling sustained improvements and the strongest final performance across all datasets.

\section{Analysis}

We now analyze SymDiag in greater depth to understand how it achieves consistent improvements in both reasoning diagnosis and repair.

\subsection{Ablation Study}
\label{sec:ablation}
\begin{figure}
    \centering
    \includegraphics[width=1\linewidth]{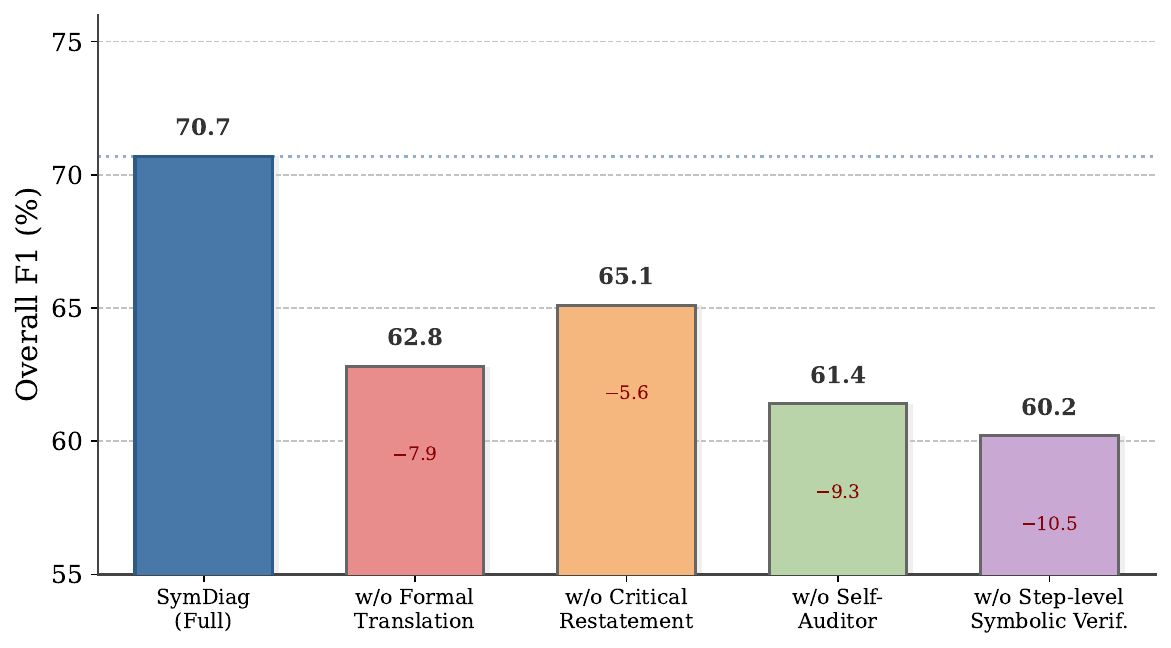}
\caption{Ablation results on overall faithfulness detection (F1).}
\label{fig:ablation_overall}
\end{figure}

Figure~\ref{fig:ablation_overall} reports a comprehensive ablation study on the overall faithfulness detection performance. Removing any single component from SymDiag leads to a consistent degradation in F1, confirming that the system’s gains do not stem from a single heuristic but from the interaction between neuro-symbolic translation, verification, and diagnosis.

The largest performance drop occurs when step-level symbolic verification is removed, indicating that explicit satisfiability and entailment checking is the primary driver of accurate failure detection. Eliminating the Self-Auditor also causes a substantial decline, as translation artifacts are more frequently misclassified as reasoning errors. Finally, ablating either the Formal Translation or Critical Restatement branch reduces performance by weakening the system’s ability to expose hidden assumptions and underspecified inferences, highlighting the importance of dual-path symbolic compilation for robust diagnosis.

\subsection{Error Attribution Analysis}

\begin{figure}[htbp]
    \centering
    \includegraphics[width=1\linewidth]{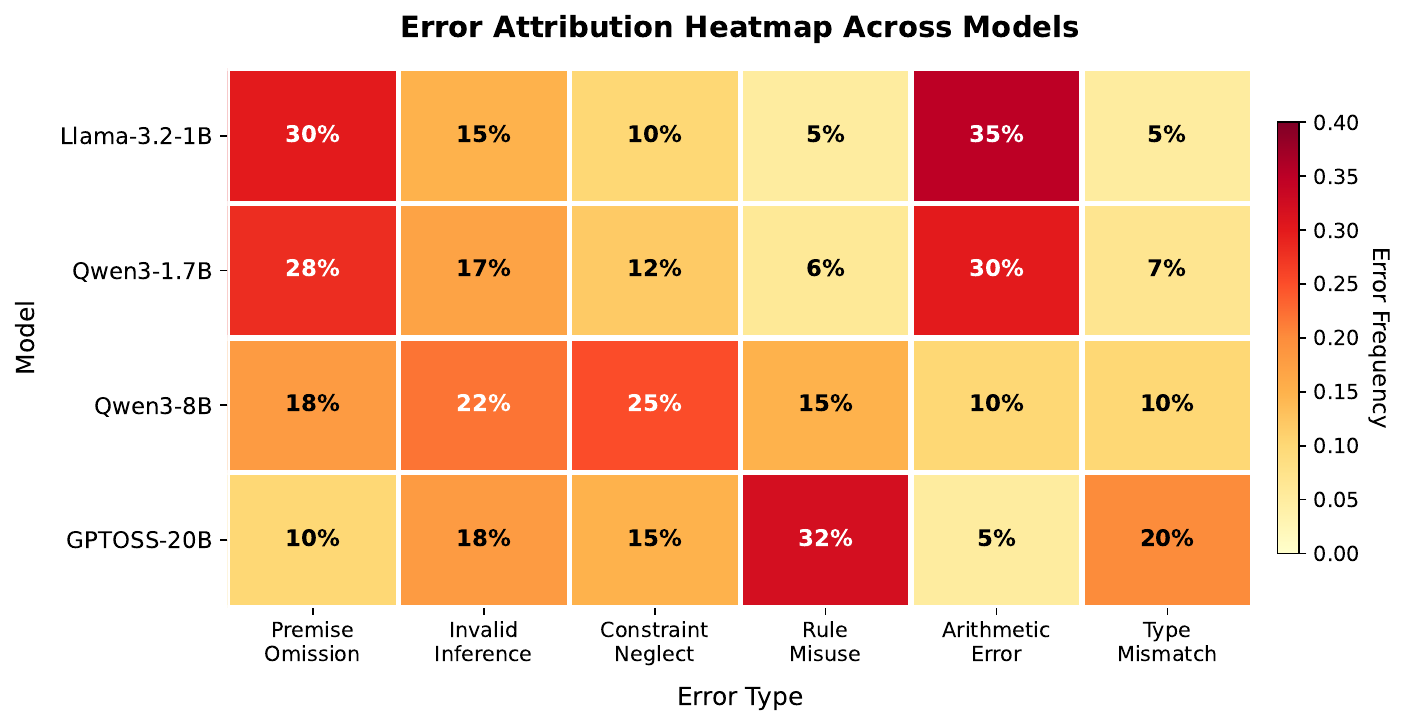}
    \caption{Normalized distribution of reasoning error types identified by SymDiag.}
    \label{fig:error_attribution_heatmap}
\end{figure}

To better understand how different models fail, we analyze the distribution of reasoning error types identified by SymDiag across four representative base models: Llama-3.2-1B, Qwen3-1.7B, Qwen3-8B, and GPTOSS-20B. For each model, we aggregate SymDiag’s diagnoses and visualize the relative frequency of each error type using a heatmap, where rows correspond to models and columns correspond to error categories defined in Section~\ref{sec:taxonomy}.

The heatmap reveals clear and systematic differences in failure modes across model scales. In summary, SymDiag reveals a clear capacity-dependent shift in reasoning failures:
(i) Small models such as Llama-3.2-1B and Qwen3-1.7B frequently commit arithmetic errors and omit necessary premises, indicating weaknesses in precise symbolic manipulation and premise tracking.
(ii) As model capacity increases, these low-level errors decrease, while higher-level structural failures become more prominent.
(iii) In particular, GPTOSS-20B exhibits relatively few arithmetic mistakes, but shows a higher incidence of hallucinated rules and type or entity mismatches, suggesting that stronger abstraction capabilities also increase the risk of unjustified generalization.

These findings suggest that reasoning supervision should be model-scale aware: smaller models benefit most from constraint enforcement and premise completion, while larger models require safeguards against high-level rule hallucination and overgeneralization.

\subsection{Effectiveness of the Self-Auditor}

\begin{figure}[htbp]
    \centering
    \includegraphics[width=1.0\linewidth]{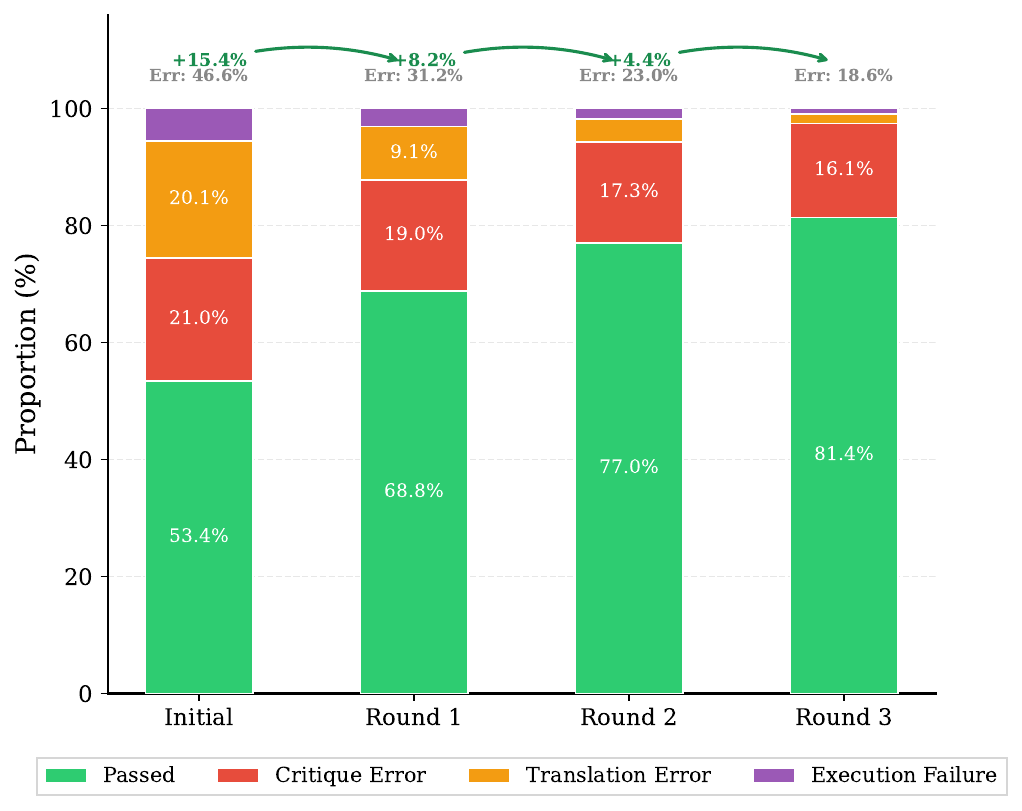}
    \caption{Progressive reduction of error types in the SymDiag pipeline through iterative Self-Auditor feedback. Translation Errors and Execution Failures are rapidly eliminated, leading to a steady increase in the overall pass rate.}
    \label{fig:error_self_auditor_repair}
\end{figure}

Figure~\ref{fig:error_self_auditor_repair} illustrates how iterative Self-Auditor feedback progressively reduces errors in the SymDiag pipeline. Before auditing, the total error rate is 46.6\%. Translation Errors(syntax-level failures in Prolog generation) are reduced most effectively, dropping from 20.1\% to near zero by Round~3, while Execution Failures similarly decline from 5.5\% to below 1\%. Correspondingly, the Passed rate increases from 53.4\% to 81.4\%, with diminishing gains across rounds. Overall, Self-Auditor substantially improves pipeline reliability, and the remaining errors are dominated by irreducible logic-level critique failures rather than translation noise.

\subsection{Generator Model Sensitivity}
\label{sec:generator_sensitivity}

\begin{figure}[t]
    \centering
    \includegraphics[width=1.0\linewidth]{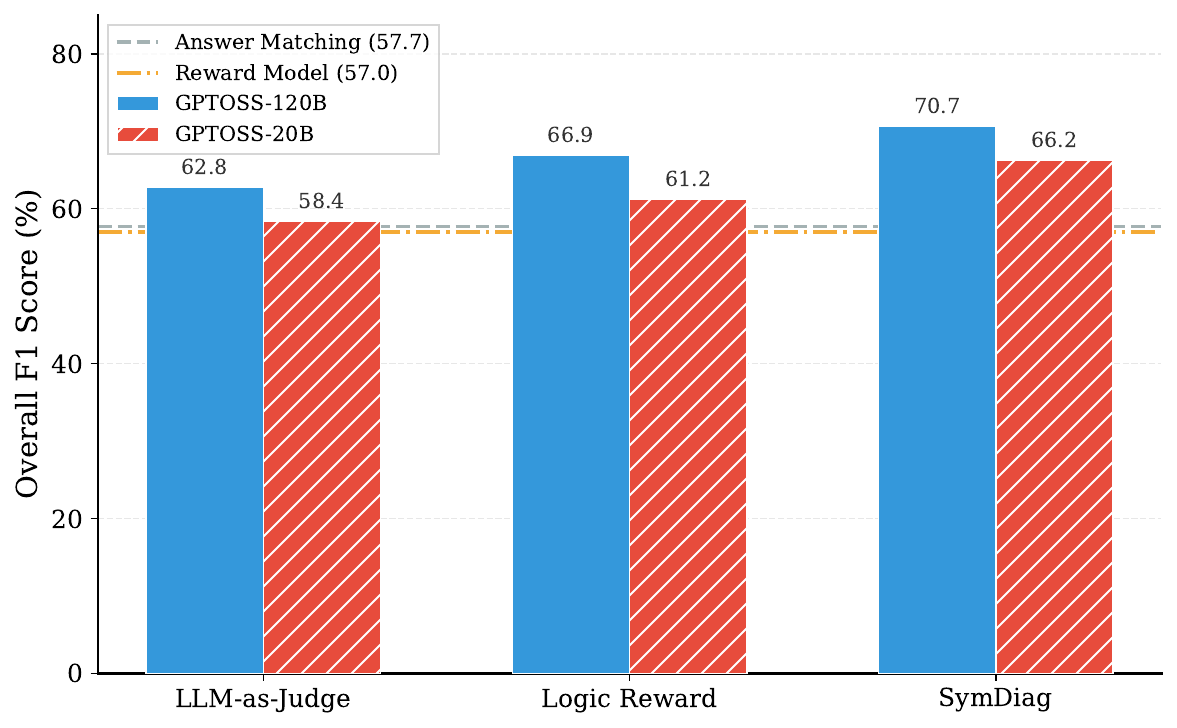}
    \caption{Effect of base model scale on overall faithfulness detection F1. Solid bars compare GPTOSS-120B and GPTOSS-20B across three methods; dashed lines indicate Answer Matching and Reward Model baselines. Larger base models consistently improve all methods, and SymDiag maintains the highest F1 under both model scales.}
    \label{fig:base_model_compare}
\end{figure}

To assess whether SymDiag's diagnostic advantage depends critically on the scale of the underlying base model, we compare all methods when using GPTOSS-120B versus GPTOSS-20B as the generator and auditor (Figure~\ref{fig:base_model_compare}). Scaling down from 120B to 20B reduces overall F1 across all methods, confirming that a more capable base model benefits every evaluation paradigm. This suggests that while stronger base models improve translation quality and diagnostic precision, the structured neuro-symbolic pipeline of SymDiag is relatively robust to model scale reduction, preserving most of its advantage over purely neural baselines.

\section{Conclusion}
We presented \textbf{SymDiag}, a neuro-symbolic framework that reframes LLM reasoning verification as an explainable failure diagnosis problem. By translating chains-of-thought into symbolic constraints and performing step-level verification, SymDiag localizes faulty inferences and produces verifiable diagnostic evidence, while a self-auditing mechanism disentangles translation artifacts from genuine reasoning errors. \textbf{SymDiag is not a better verifier or a better reward, but an explainable neuro-symbolic diagnostic system that localizes, attributes, and repairs reasoning failures with verifiable evidence across domains.} Experiments across mathematical, logical, scientific, and general reasoning tasks show that SymDiag outperforms outcome-based and judge-based baselines in detecting unfaithful reasoning and provides substantially more effective feedback for multi-round reasoning repair, especially for smaller models. Future work includes extending the symbolic backend to stronger or hybrid verifiers, improving robustness to underspecified language, and leveraging diagnostic evidence to train diagnosis-aware reward models and reasoning supervisors with efficient and meta-learning paradigms~\cite{guan2025learning,guan2025optimizer,guan2025meta}.

\section{Acknowledgments}
This work is supported by Zhongguancun Academy Project No.02012501, in part by the National Natural Science Foundation of China (NSFC) Grant 62436009.

\bibliographystyle{ACM-Reference-Format}
\balance
\bibliography{sample-base}

\appendix

\section{Error Taxonomy}
\label{sec:taxonomy}

We use a domain-agnostic taxonomy designed to cover common LLM failure modes across math, logic, science, and general reasoning:

\begin{itemize}
    \item \textbf{Premise Omission / Missing Assumption:} step claim underivable without additional assumptions;
    \item \textbf{Invalid Inference:} non-entailing transformation (e.g., illicit equivalence, quantifier shift);
    \item \textbf{Constraint / Boundary Neglect:} ignoring domain restrictions, case splits, sign constraints, unit constraints;
    \item \textbf{Rule Misuse / Hallucinated Rule:} applying an inapplicable theorem or principle;
    \item \textbf{Arithmetic / Algebra Error:} verified mismatch in computed equalities/inequalities;
    \item \textbf{Type / Entity Mismatch:} inconsistent predicate arity/type or category confusion;
    \item \textbf{TranslationError:} solver failure attributable to compilation rather than reasoning.
\end{itemize}

\end{document}